\documentclass[10pt,english,final,a4paper]{article}

\usepackage[usenames,dvipsnames]{color}
\usepackage{hyperref} % Hyperlinks within and outside the document
\hypersetup{
unicode=false,          % non-Latin characters in Acrobat’s bookmarks
pdfauthor={JuanPi Carbajal},%
pdftitle={MBM},%
colorlinks=true,       % false: boxed links; true: colored links
linkcolor=OliveGreen,          % color of internal links
citecolor=Sepia,        % color of links to bibliography
filecolor=magenta,      % color of file links
urlcolor=NavyBlue,          % color of external links
}

\usepackage[numbers, colon, sort&compress  ]{natbib}
\usepackage{nicefrac}

\usepackage[utf8]{inputenc}
\usepackage[T1]{fontenc}
\usepackage{textcomp}
\usepackage{lmodern} % German related symbols

\usepackage{babel}			 %la opcin spanish de arriba la toma como global
\usepackage{graphicx}		%incluye graficos, la opcin esta global, dvips
\usepackage[small,labelfont=bf]{caption} %To change the appearance of captions, use the caption package. E.g. to make all caption labels small
\usepackage[cdot,squaren,derived]{SIunits}
\usepackage{subcaption}

\usepackage{booktabs}

\usepackage{amsmath}    %formulas matematicas lindas
\usepackage{amsfonts}
\usepackage{amssymb}
\usepackage{amsthm}
\usepackage{thmtools}
\usepackage{authblk}
\usepackage{bbold}

\usepackage{lineno} % Line numbers

\definecolor{mygreen}{rgb}{0,0.6,0}
\definecolor{mygray}{rgb}{0.5,0.5,0.5}
\definecolor{mymauve}{rgb}{0.58,0,0.82}
\usepackage{listings}
\newcommand{\ud}{\mathrm{d}}

\newcommand{\definter}[4]{\int_{#1}^{#2}\! {#3}\, \ud {#4}}

\newcommand{\bm}[1]{\boldsymbol{#1}}

\newcommand{\vBQ}{\boldsymbol{\Theta}}

\DeclareMathOperator*{\dist}{dist}

\graphicspath{ {figs/} }

\defcitealias{Octave}{GNU Octave}
\defcitealias{Inkscape}{Inkscape}
\defcitealias{Sage}{Sage}
\defcitealias{LIACC}{LIACC datasets}

\begin{document}

\title{Memristor models for machine learning}
\author{Juan Pablo Carbajal}
\author{Joni Dambre}
\author{Michiel Hermans}
\author{Benjamin Schrauwen}
\affil{Department of Electronics and Information Systems, Ghent University, Belgium\\
{\tt\small \{juanpablo.carbajal, michiel.hermans, benjamin.schrauwen, joni.dambre\}@ugent.be}
}

\date{\today}

\maketitle

%\linenumbers

\begin{abstract}
In the quest for alternatives to traditional CMOS, it is being suggested that digital computing efficiency and power can be improved by matching the precision to the application.
Many applications do not need the high precision that is being used today. 
In particular, large gains in area- and power efficiency could be achieved by dedicated analog realizations of approximate computing engines.
In this work, we explore the use of memristor networks for analog {\em approximate} computation, based on a machine learning framework called {\em reservoir computing}.
Most experimental investigations on the dynamics of memristors focus on their nonvolatile behavior. 
Hence, the volatility that is present in the developed technologies is usually unwanted and it is not included in simulation models.
In contrast, in reservoir computing, volatility is not only desirable but necessary. 
Therefore, in this work, we propose two different ways to incorporate it into memristor simulation models. 
The first is an extension of Strukov's model and the second is an equivalent Wiener model approximation. 
We analyze and compare the dynamical properties of these models and discuss their implications for the memory and the nonlinear processing capacity of memristor networks.
Our results indicate that device variability, increasingly causing problems in traditional computer design, is an asset in the context of reservoir computing.
We conclude that, although both models could lead to useful memristor based reservoir computing systems, their computational performance will differ. 
Therefore, experimental modeling research is required for the development of accurate volatile memristor models.
\end{abstract}

\section{Introduction}
As the scaling of the traditional MOSFET transistor is reaching its physical limits, alternative building blocks of future generation computers are being investigated.
Most of this research is focused on producing controllable switch-like behavior, but over the last decade analog computation has also enjoyed a revival, partly due to the increasing success of neuromorphic devices.
In particular, memristor networks have been used for computational purposes, mostly using structured topologies, since their fabrication exploits technology developed for the assembly of cross-bar arrays of transistors~\citep[][ section 5]{Indiveri2013,Kuzum2013}. 
In most of these applications, memristors are used as programmable synaptic weights.
However, at the nanoscale, even highly controlled processing flows cause relatively large variations in the device parameters.
For this reason, computing systems built with such components must embrace this variability~\citep{Indiveri2013}. 
Increasingly, it is being suggested that efficiency and power gains can be achieved in digital computers by exploiting the fact that many applications do not need the high precision that is being used today. 
In this new research field of {\em approximate computing}, digital, low-precision neural network accelerators are currently being evaluated ~\citep{Esmaeilzadeh2012, Samadi2013}.

Reservoir computing (RC) is a supervised learning framework rooted in recurrent neural network research (seminal work in~\citep{Maass2002, Jaeger2004}, recent developments in~\citep{Maass10,Lukosevicius2012}).
It was originally applied to simulated neural networks in discrete time and can be implemented in open loop operation~\citep{Triefenbach10, Buteneers2013, Ongenae2013, Alessandro2013, Sillin2013} or in feedback systems~\citep{Reinhart2012, Fiers2014, wyffels2014}.
RC can also be used as a leverage to directly exploit the analog computation that naturally occurs in physical dynamical systems in a robust way, i.e.\ without the need to control all system parameters with high precision.  
Recently, this approach ({\em physical} RC or PRC) has been demonstrated (in simulation or experimentally) using several physical substrates. 
These include a water tank~\citep{Fernando2003}, tensegrity structures~\citep{Caluwaerts2013}, opto-electronic and nano-photonic devices~\citep{Vandoorne2014,Paquot2012,Larger2012,Fiers2014} and resistive switch networks~\citep{Sillin2013}. 

The computation in RC is based on the observed responses of a set of (interacting input-driven nonlinear) dynamical systems. 
At each point in time, the $n$ observed responses are linearly combined, in order to optimally approximates a desired output.
As is common in supervised machine learning approaches, the weights in this linear combination are optimized for a set of representative examples of the desired input-output behavior (i.e., the {\em task}).
The system's responses to each of the input signals are sampled and the $m$ samples recorded into a large $m \times n$ matrix called the {\em design matrix} $\vBQ$, which is usually used in a common linear regression setup:

\begin{equation}
\bm{\hat{y}} = \vBQ \bm{w} \quad \text{minimizing} \quad \dist(y,\hat{y}),\label{eq:RC}
\end{equation}
\noindent where $\hat{y}$ is the (sampled) desired output. 
The readout vector $\bm{w}$ has $n$ components, one for each observed state signal, and remains fixed after optimization ({\em training}).
The aim is to optimize the readout weights in such a way that the system {\em generalizes} well to new data that has not been used for training.
Note that a task is comprised of more than one desired output.
The task can be described by comprehension, as in delay tasks or filtering tasks, in which case the desired outputs used to optimize ({\em train}) the readout weights are sampled from the set and the amount of training data is potentially unbounded (also known as synthetic data).
However, as in many machine learning techniques, the task is such that no analytical or algorithmic solution is available. 
In this case, tasks is known only through a finite number of input-output examples.

In order to obtain useful results with PRC, the system's responses to the input signals must meet several requirements.
First, reservoir computing strongly relies on a property of the input-output relation of the system generating the responses, namely \textit{fading memory} (also known as the echo state property).
It is present when the output depends on previous inputs of the system in a decaying way, such that perturbations far in the past do not affect the current state.
In this way, the functional relationship between the response of the system and the inputs is localized in time, providing a notion of "memory"~\citep{Hermans2010}.
Another role of fading memory is to avoid high sensitivity to initial conditions which, if present, could obscure the relation between inputs and outputs~\citep[see][and references therein]{Manjunath13}.

Secondly, analog computation, and in particular RC, exploits the natural behavior of the device (the computer) to simulate a primary system.
The class of computable primary systems is strictly related to the class of systems used as computer, e.g.\ both systems are modeled with the same class of differential equations~\citep[see "General-Purpose Analog Computation" in][]{MacLennan2012}.
In RC this is reflected in the properties of the generated design matrix.

In general $m \gg n$ and the linear problem in Eq.~\eqref{eq:RC} is rank-deficient or ill-posed~\citep{Hansen1998}.
The suitability of a dynamical system for reservoir computing depends on the extent to which input signals affect the system responses.  
In particular, the rank and range of the design matrix play a crucial role.
Intuitively, the numerical rank $r$ of $\vBQ$ characterizes the power to reproduce features of the desired output.
Indeed, in a geometrical interpretation, the responses in the design matrix span an $r$-dimensional subspace of all possible features~\citep{Dambre2012}.
Hence, for large ranks, one expects that the features required for a specific task are present within this subspace.
The affinity between the task and the design matrix depends on the linear span of its singular vectors (the range of $\vBQ$).
Highly complicated responses (e.g. characterized by randomness or by the chaotic dynamics of the underlying autonomous system), produce design matrices with high ranks but with extremely varied ranges.
These matrices do not generalize well to unseen data, which leads to poor task performance.
Moreover, if the responses do not significantly depend on the input history, they will not be very useful as features for obtaining a desired input-output relationship.

Whether or not a system can meet these requirements is for a large part determined by the system's inherent dynamics. 
Hence, to gauge what can be computed with a given system, it is crucial to understand its behavior.
This allows us to propose encodings for the inputs and parameter values that optimize computation. 

In this paper we revisit the implementation of RC using volatile resistive switches (for short we will call them memristors or memristive devices, despite the ongoing controversy~\citep{Vongehr2012}). 
This is a novel application of memristive devices as processing units rather than mere synaptic weights.
Recent work has reported large random networks of memristors that can be fabricated with relative ease~\citep{Avizienis2012}.
Their dynamical behavior indicates that they could be used as reservoirs~\citep{Stieg2012,Sillin2013}.

\begin{figure}[htpb]
\centering
\includegraphics[width=0.5\textwidth]{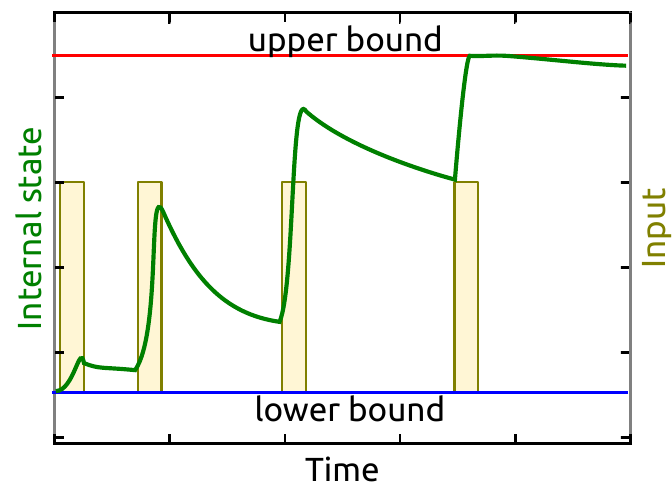}
\caption{State dependent diffusion. Schematic reproduction of Fig. 1 of~\citep{Ohno2011}.
\label{fig:state_decay}}
\end{figure}

Most experimental investigations focus on the nonvolatile dynamics of memristors. 
Interestingly, volatility is present in the developed technologies~\citep[e.g. Fig. 1 of][see Fig.~\ref{fig:state_decay}]{Ohno2011}.
It can be observed that the conductance of the memristive device under study decays when there is no input, unless the conductive silver filament is fully formed.
Additionally, the rate of decay depends on the stage of development of the filament: slow decay for incipient filaments and for those that are close to being fully formed.
Fig.~\ref{fig:state_decay} sketches the state dependent decay rate. 
This behavior indicates a state dependent volatility with a maximum at some intermediate state of formation of the silver filament.

Due to the focus on ultra-dense storage (ReRAMs) applications, the current trend is to engineer volatility out of the devices.
Consequently, reports of numerical models suitable for the simulation of volatile memristor networks are scarce~\citep[][ section 3.2]{Kuzum2013}.
The available mathematical models of memristors are adequate for the study of nonvolatile devices, i.e. having internal states without (or with negligible) autonomous dynamics.

To adequately study the potential use of memristor networks for RC, which requires fading memory, a deeper study of volatility is in order.
For example, when working with memristors, volatility prevents saturation.
This is key for RC, since a saturated memristor becomes a linear resistor that only scales the input.
In this work we present a first study on models of memristors that focus on volatility and their dynamical response.
These models are studied to provide a conceptual basis for the study of networks of these devices and their use in RC.
The article begins with a short recapitulation of general models of memristive systems.
Section~\ref{sec:hp} introduces the model proposed by Strukov~\citep{Strukov08} and its current driven solution, and proposes an equivalent nonvolatile Wiener model.
Section~\ref{sec:fading} presents a simple modification of each model that introduces linear volatility.
We focus on the production of harmonics and delays in the steady state response of these models, in the vein of the harmonic balance method.
These aspects are important when designing the encoding to use for the inputs representing values in datasets.
Section~\ref{sec:series} describes RC using memristors connected in series and two simple application examples are given.
Subsequent sections discuss the results and indicate future directions of research.

\section{Memristive systems\label{sec:mems}}
Memristive systems are devices that can be modeled with a nonautonomous (input-driven) dynamical system of the form,

\begin{align}
\bm{\dot{x}} = F(\bm{x},u), \label{eq:dynamics} \\
y = H(\bm{x},u) u \label{eq:output}
\end{align}

\noindent where $\bm{x}$ is a vector of internal states, $y$ a measured scalar quantity and $u$ a scalar magnitude controlling the system. 
Due to its origins in electronics~\citep{Chua1971}, the pair $(y,u)$ is usually voltage-current or current-voltage.
In the first case ("current controlled" system) the function $H(\bm{x},u)$ is the called "memristance", and it is called "memductance" in the second case ("voltage controlled").
We use a general denomination for $H(\bm{x},u)$ and call it \emph{output function}.
The output function is assumed to be continuous, bounded and of constant sign, leading to the zero-crossing property: $u = 0 \Rightarrow y = 0$.
%It has been shown that this property is a modeling deficiency for redox-based resistive switches, and extended memristive system was proposed~\citep{Valov2013}.
It has been shown that this property is a modeling deficiency for redox-based resistive switches~\citep{Valov2013}, and an extended version of equation~\eqref{eq:output} was proposed, namely

\begin{equation}
y = H(\bm{x},u) u + h(\bm{x}).\label{eq:extmem}
\end{equation}

\noindent Where $h(\bm{x})$ is an autonomous output given by an internal battery in the system.
This term, in general, removes the zero-crossing property making these systems incompatible with the original definition of memristor.
In this work, we will not study these extended systems.

%In the cases where $u$ is the current flowing through the device, the system is denominated "current controlled" and the function $H(\bm{x},u)$ is the called "memristance".
%If $u$ is the voltage across the terminals of the device, the system is denominated "voltage controlled" and the function $H(\bm{x},u)$ is the called "memductance".
%We use a general denomination for $H(\bm{x},u)$ and call it \emph{output function}.
%The output function is assumed to be continuous and of constant sign.
%Furthermore, this function is bounded for all values of its arguments, i.e. $h \leq H(\bm{x},u) \leq H$.
%The latter assumption leads to one important characteristic of these models (zero-crossing property): $u = 0 \Rightarrow y = 0$.
%It has been shown that this property is a modeling deficiency for redox-based resistive switches~\citep{Valov2013}, and an extended version of equation~\eqref{eq:output} was proposed, namely

%\begin{equation}
%y = H(\bm{x},u) u + h(\bm{x}).\label{eq:extmem}
%\end{equation}

%\noindent Where $h(\bm{x})$ is an autonomous output given by an internal battery in the system.
%This term, in general, removes the zero-crossing property making these systems incompatible with the original definition of memristor.
%In this work, we will not study these extended systems.

\subsection{Memristor models}
\label{sec:hp}
One of the most popular models of memristive behavior consists of a single internal state with linear dynamics and a piece-wise linear output function~\citep{Strukov08},

\begin{align}
&\dot{x} = \mu I(t),\label{eq:ldrift}\\
&H(x) = \begin{cases}R & \text{if } x < 0 \\
                         R - (R-r)x & \text{if } 0 \leq x \leq 1\\
                         r & \text{if } x > 1\end{cases},\\
&V(t) = H(x) I(t),
\end{align}

\noindent where $I,V$ are the current and voltage, respectively and $\mu$ is a parameter related to the geometry and the physical properties of the material realizing the memristor. 
In particular it is proportional to the average ionic mobility.
$H(x)$ plays the role of a state dependent resistance.
Note that the value of $x$ can grow unboundedly while $H(x)$ remains bounded.
For this reason some authors consider the model valid only for $x \in [0,1]$\citep[Eq. 31]{Strukov08, Biolek2012}.

A modified version of this model applies a windowing function to the dynamics of the internal state, effectively bounding it~\citep{Strukov08, Oskoee2011}

\begin{gather}
\begin{aligned}
\dot{x} &= \mu x (1-x)I(t),\\
H(x) &= R - \Delta r x,\\
V(t) &= H(x) I(t).
\end{aligned}\label{eq:nldrift}
\end{gather}

\noindent The parameter $\Delta r = R-r$ depends on the bounds of the output function.
The internal state is driven by a Bernoulli equation, hence we can obtain an explicit input-output relation for the current controlled memristor,

\begin{equation}
V(t) = \left[R -  \Delta r\left(1 + \frac{1-x_0}{x_0}e^{-\mu q(t)}\right)^{-1}\right]I(t),\label{eq:Icontrol}
\end{equation}

\noindent where $q(t) = \definter{0}{t}{I(\tau)}{\tau}$. 

Noteworthy, relation~\eqref{eq:Icontrol}, corresponding to the nonlinear internal state dynamics~\eqref{eq:nldrift} and linear output function, is indistinguishable from the relation corresponding to a system with linear internal state dynamics and sigmoidal output function, i.e. a Wiener model defined as

\begin{gather}
\begin{aligned}
\dot{q} &= \mu I(t),\\
H(q;x_0) &= R - \Delta r \left(1 + \frac{1-x_0}{x_0}e^{-q(t)}\right)^{-1} \\
         &= R - \frac{\Delta r}{2}\left[\tanh\left(\frac{q(t)}{2} +C(x_0)\right)+1\right].\\
\end{aligned}\label{eq:linmodel}
\end{gather}

\noindent Note that this model is a smoothed version of the original Strukov model.
The fact that the nonlinear model can be seen as a Wiener system imposes strong restrictions to the type of processing the system will be able to perform on the input signals.
For example, its memory function will be restricted to exponential decays, as in linear systems.
As will be discussed next, this equivalence is broken when an autonomous term is added to the dynamics.
However, in Section~\ref{sec:wiener} we will establish an approximated equivalence for small amplitude input signals, indicating that strong amplitudes will be needed to depart form a linear memory behavior.

\section{Fading memory}
\label{sec:fading}
We propose a modification of these models that includes an autonomous term,
\begin{equation}
\dot{x} = F(x,u) + D(x)\label{eq:extdyn}
\end{equation}
\noindent where for all $x$, we have that $F(x,0) = 0$ and $D(x) < 0$.
In the sections that follow we will restrict ourselves to the case when $D(x)$ is a linear function.
Note the difference with Eq.~\eqref{eq:extmem}. 
Since we are adding \emph{diffusive} dynamics to the internal state, it does not change the zero-crossing property of the memristive system.
This modification of the model in the context of resistance switching, arising from charge transport, implies that there is a flow of carriers that goes out of the electrical circuit of the device (e.g. additional sink) and it is not explicitly modeled.
For example, in the hydraulic memristor studied in~\citep{Biolek2012}, the decay of the internal state is realized when the storage tank has a leak, that reduces the hydraulic resistance.
%Note that by doing this the methods to convert between current and voltage controlled devices described by ~\citeauthor{Biolek2012} have to be extended, since these volatile models are not time invariant.
Note that by doing this the methods to convert between current and voltage controlled devices described by Biolek~\citep{Biolek2012} have to be extended, since these volatile models are not time invariant.
As said before, the equivalence of the models~\eqref{eq:nldrift} and~\eqref{eq:linmodel} is lost.

\subsection{Nonlinear dynamics}
In the case of the nonlinear dynamics model~\eqref{eq:nldrift} we write,
\begin{equation}
\dot{x} = \mu x (1-x)I(t) - \lambda x = (\mu I(t)-\lambda)x - \mu I(t)x^2.\label{eq:diff_nldrift}
\end{equation}
\noindent This equation can again be identified with the Bernoulli equation and the solution is (see Supplementary data for derivation)
\begin{align}
x(t) &= \left(1 + \frac{1-x_0}{x_0}F(t)^{-1} + \lambda F(t)^{-1}\definter{0}{t}{F(z)}{z}\right)^{-1}\\
F(t) &= e^{\definter{0}{t}{\mu I(w)}{w}-\lambda t}.\label{eq:gen_sol}
\end{align}
\noindent Comparing with Eq.~\eqref{eq:Icontrol}, we see that $\lambda$ introduces an exponentially decaying factor (linearly detrending the integral of the input) and adds the integral term.
In the mathematical description that follows, we investigate the harmonics and delay generation in the steady response of the system.
The presence of harmonics directly impacts the maximum rank that the RC design matrix can achieve.
Similarly, as any delayed sinusoidal signal can be expressed as a superposition of the undelayed signal plus a cosine component of the same frequency, delays can also increase the maximum rank of the design matrix.
We begin by expressing the convergence of the mean response (obtainable directly from the fixed point of Eq.~\eqref{eq:diff_nldrift}) and continue with the response generated by periodic input signals.
We explicitly write the delay and harmonic generation in terms of interactions between parameters of the input signal and the system's parameters.

Using the first mean value theorem for integration when the input has mean value $\mu\bar{I} = m$ results in, 
\begin{equation}
\bar{x}(t) = \left(1 + \frac{\lambda}{m- \lambda} +\frac{1-x_0}{x_0}e^{\left(\lambda-m\right)t}\right)^{-1},
\end{equation}
\noindent which is nonzero in the long run only if $m>\lambda$. This average evolution has a pole at 
\begin{equation}
t_c = \frac{1}{\lambda - m} \ln \left[\left(\frac{m}{\lambda-m}\right)\left(\frac{x_0}{1-x_0} \right)\right].
\end{equation}
Taking $t = t_c + \tau$ the convergence to the steady state solution of Eq.~\eqref{eq:diff_nldrift} is given by 
\begin{equation}
\bar{x}(t) = \frac{m-\lambda}{m}\left(1 - e^{(\lambda -m)\tau}\right)^{-1},\label{eq:avg_converg}
\end{equation}
\noindent which shows that all initial conditions converge to the same mean but their convergence time increases with decreasing initial conditions.
In other words, a memristor with high resistance will take longer to converge to the steady state given by the mean value of the input.
However, the timing properties of Eq.~\eqref{eq:gen_sol} do not depend on the initial condition. 
Therefore delays (and fading memory) depend only on the interaction between the input and the memristor parameters.

This interaction is crucial to determine how the system reacts to different inputs signals.
We characterize two easily controllable properties of the inputs, namely, their mean value and their deviations from it.
Deviations from the mean value can be seen as small amplitude oscillations.
Therefore, to evince the interaction between system's parameters and the input signal, the latter is decomposed in its mean value and a zero-mean waveform:

\begin{align}
\mu I(t) &= m + \gamma(t), \\
\mu \definter{0}{t}{I(\tau)}{\tau} &= m t +  \phi(t) + \phi_0, 
\end{align}

\noindent where $\phi(t)$ is the primitive of $\gamma(t)$ and the initial values were all aggregated in the constant $\phi_0$.
We assume that both functions are bounded and proceed to calculate the harmonics generation and delay between input and output.
For this family of inputs we get,

\begin{equation}
x(t)^{-1} = 1 + e^{-\phi_0} e^{-\phi(t)}\frac{1-x_0}{x_0}e^{(\lambda - m)t} + \lambda \definter{0}{t}{e^{\phi(z)-\phi(t)}e^{(m-\lambda)(z-t)}}{z}.\label{eq:decomp}
\end{equation}

\noindent Assuming that $m = \lambda + \epsilon$ with $\epsilon > 0$, the second term vanishes in the long run.
Additionally, if we assume small bounds for $\phi(t)$, i.e. $\vert \phi(z) - \phi(t)\vert$ small, we can expand the integrand obtaining,

\begin{equation}
\definter{0}{t}{e^{\phi(z)-\phi(t)}e^{-\epsilon(z-t)}}{z} = \frac{1-e^{-\epsilon t}}{\epsilon} + \sum_{n=1}^\infty \frac{1}{n!}\definter{0}{t}{\left(\phi(z)-\phi(t)\right)^n e^{\epsilon(z-t)}}{z}.\label{eq:taylor_resp}
\end{equation}

\noindent This expansion makes evident that the response of the system can be seen as the infinite superposition of the response of a linear system to all the homogeneous monomials in $\lbrace\phi(z),\phi(t)\rbrace$.
Inserting this expansion back into Eq.~\eqref{eq:decomp}, keeping terms of first order and removing vanishing terms we obtain,

\begin{equation}
x(t)^{-1} = \frac{\epsilon + \lambda}{\epsilon} + \lambda\definter{0}{t}{\left(\phi(z)-\phi(t)\right) e^{\epsilon(z-t)}}{z}.
\end{equation}

As a prototype of zero mean waveforms we take $\gamma(t) = \alpha\sin(\omega t)$ and proceed to specify the previous equations on inputs of this kind (this will provide the harmonic response of the memristor).
Introducing this input into the previous equation we get,

\begin{align}
x_\omega(t) &= \left(\frac{\epsilon + \lambda}{\epsilon} - \frac{\lambda}{\epsilon}\frac{\alpha}{\sqrt{\omega^2 + \epsilon^2}}\sin(\omega t - \varphi)\right)^{-1}\label{eq:freq_response0}\\
            &= \frac{\epsilon}{m}\left(1 - \frac{\lambda}{m\omega}\alpha\sin(\varphi)\sin(\omega t - \varphi)\right)^{-1},\label{eq:freq_response}
\end{align}

\noindent where vanishing terms were removed (note $ m = \epsilon + \lambda$) and the delay is given by

\begin{equation}
\sin(\varphi) = \frac{\omega}{\sqrt{\omega^2 + \epsilon^2}}.\label{eq:delay}
\end{equation}

Note that, as long as

\begin{equation}
\alpha < \frac{m\sqrt{\omega^2 + \epsilon^2}}{\lambda},
\end{equation}

\noindent we can further expand Eq.~\eqref{eq:freq_response}:

\begin{equation}
x_\omega(t) = \frac{\epsilon}{m}\left[1 + \sum_{n=1}^\infty \left(\frac{\lambda}{m\omega}\right)^n\alpha^n\big(\sin(\varphi)\sin(\omega t - \varphi)\big)^n\right].\label{eq:freq_response2}
\end{equation}

\noindent This result is a good approximation for small amplitudes of the inputs.
For larger amplitudes, higher order terms in Eq.~\eqref{eq:taylor_resp} are not negligible and harmonics of the input frequency increase their contribution.

The voltage drop across the memristor is obtained using Eq.~\eqref{eq:nldrift} by multiplication with the input current and has three terms. 
The first term is the voltage drop across an effective resistor, and has the same frequency as the input:
\begin{equation}
\frac{r m + \Delta r \lambda}{\mu} \left(1 + \frac{\alpha}{m}\sin(\omega t)\right).
\end{equation}

\noindent The second term is obtained by multiplication of the oscillatory part of $x_\omega(t)$ with the mean value of the input:

\begin{equation}
- \frac{\Delta r (m - \lambda)}{\mu} \sum_{n=1}^\infty \left(\frac{\lambda}{m\omega}\right)^n\alpha^n\big(\sin(\varphi)\sin(\omega t - \varphi)\big)^n.
\end{equation}

\noindent The third term is formed by the product of the oscillatory parts of the input and $x_\omega(t)$,

\begin{equation}
- \frac{\Delta r (m - \lambda)}{m\mu} \sum_{n=1}^\infty \left(\frac{\lambda}{m\omega}\right)^n\alpha^{n+1}\big(\sin(\varphi)\sin(\omega t -\varphi)\big)^n \sin(\omega t).
\end{equation}

The voltage drop is the sum of all these terms. Taking $n=1$ and using trigonometric identities, the coefficients of the sine ($a_\omega$) and cosine ($b_\omega$) are:
\begin{align}
a_\omega &= \left(\frac{\alpha}{m}\right)\left(\frac{\Delta r \lambda}{\mu}\sin^2(\varphi) + r\right) = 
\left(\frac{\alpha}{m}\right)\left(\frac{\Delta r \lambda}{\mu}\frac{\omega^2}{\omega^2+(m-\lambda)^2} + r\right),\label{eq:sine_w}\\
b_\omega &= \left(\frac{\alpha}{m}\right)\left(\frac{\Delta r \lambda}{\mu}\cos(\varphi)\sin(\varphi)\right) = \left(\frac{\alpha}{m}\right)\left(\frac{\Delta r \lambda}{\mu}\frac{\omega(m-\lambda)}{\omega^2+(m-\lambda)^2}\right).\label{eq:cosine_w}
\end{align}
\noindent For the first harmonic $2\omega$ we obtain:
\begin{align}
a_{2\omega} &= \frac{1}{2}\left(\frac{\alpha}{m}\right)^2\left(\frac{\Delta r \lambda}{\mu}\cos(\varphi)\sin(\varphi)\right) = \frac{1}{2}\left(\frac{\alpha}{m}\right)b_\omega,\label{eq:sine_2w}\\
b_{2\omega} &= \frac{1}{2}\left(\frac{\alpha}{m}\right)^2\left(\frac{\Delta r \lambda}{\mu}\cos^2(\varphi)\right) =
\frac{1}{2}\left(\frac{\alpha}{m}\right)^2\left(\frac{\Delta r \lambda}{\mu}\frac{(m-\lambda)^2}{\omega^2+(m-\lambda)^2}\right).\label{eq:cosine_2w}
\end{align}
The amplitude of the cosine contribution at the input frequency (and, as a consequence, the delay) is modulated by the ratio between the oscillation amplitudes and the mean value. 
This contribution is shown in the top-right panel of Figure~\ref{fig:series}.
Additionally it depends on the frequency of the input signal and the difference $m-\lambda$. 
As was anticipated, the signal parameters play a mayor role in the response of the system and this can be used to propose "meaningful" encodings.

%The first order of the voltage drop is
%\begin{align}
%&V(t) = \left(r + \Delta r\frac{\lambda}{\mu I_0}\right) \big( I_0  + I_w\sin(\omega t)\big) - \nonumber\\
%&- \Delta r \frac{\lambda}{\omega}\left(1 - \frac{\lambda}{\mu I_0}\right)I_w\sin(\varphi)\left\lbrace \phantom{\frac{I_w}{I_0}}\right.\nonumber \\
%&\cos(\varphi)\sin(\omega t) - \sin(\varphi)\cos(\omega t) -\nonumber \\
%&\left. -\frac{I_w}{2I_0}\big[ \sin(\varphi)\sin(2\omega t) + \cos (\varphi)\big(\cos(2\omega t) -1\big)\big]\right\rbrace.\label{eq:Vdrop}
%\end{align}
%\noindent Where we replaced $m = \mu I_0$ and $\alpha = \mu I_w$ and expanded $\sin(\omega t - \varphi)$.

%The ratio between the cosine components and the sine components at frequencies $\omega$ and $2\omega$, are given by the coefficients,

%\begin{align}
%% \frac{\mu I_0 - \lambda}{\omega}\left(\frac{r\mu I_0}{\Delta r \lambda \sin^2(\varphi)} + 1 \right)^{-1}
%c_\omega =& \frac{\left(\frac{\mu I_0 - \lambda}{\omega}\right)}{1 + \left(\frac{r}{\Delta r}\right)\left(\frac{\mu I_0}{\lambda}\right) \left[1 + \left(\frac{\mu I_0 - \lambda}{\omega}\right)^2\right]},\label{eq:cosine_w}\\
%%\cot(\varphi) = 
%c_{2\omega} =& \quad \frac{\mu I_0 - \lambda}{\omega}.
%\end{align} 

\subsection{Wiener model}
\label{sec:wiener}
In the case of the Wiener model~\eqref{eq:linmodel} we write,

\begin{equation}
\dot{z} = \mu I(t) - \lambda_\ell (z - z_s), \quad z_s < 0.
\end{equation}

\noindent Where $z_s$ is chosen such that the output function saturates to $R$ when there are no inputs.
Note that when $z_s \neq 0$, the diffusive term is not strictly negative as requested in~\eqref{eq:extdyn}, the role of $z_s$ is to prevent numerical overflows.
The solution is readily obtained by integration:

\begin{equation}
z(t) = \mu\definter{0}{t}{I(\tau) e^{\lambda_\ell (\tau-t)}}{\tau} + (z_0 - z_s)e^{-\lambda_\ell t} + z_s
\end{equation}

\noindent and the timing properties are independent of the input characteristics. 
This makes the calculation of the harmonic response of the system trivial, so we use inputs of the form,

\begin{equation}
\mu I(t) = m + \sum_{i=1}^N \alpha_i\sin(\omega_i t).
\end{equation}
\noindent Replacing in the corresponding equations we obtain:

\begin{align}
z(t) &= \overbrace{\frac{m}{\lambda_\ell}+z_s}^{\epsilon_\ell} + \overbrace{\sum_{i=1}^N \frac{\alpha_i}{\sqrt{\omega_i^2 + \lambda_\ell^2}}\sin(\omega_i t - \varphi_i)}^{q(t)},\\
\sin (\varphi_i) &= \frac{\omega_i}{\sqrt{\omega_i^2 + \lambda_\ell^2}}. 
\end{align}
\noindent Since this is the response of a linear system, it generates no harmonics.
Higher frequencies are introduced when the response becomes the argument of the sigmoid function in Eq.~\eqref{eq:linmodel}.

\begin{multline}
x(t)^{-1} = 1 + \frac{1-x_0}{x_0}e^{-\epsilon_\ell}e^{-q(t)} = \\
1 + \frac{1-x_0}{x_0}e^{-\epsilon_\ell}\left(1 + \sum_{n=1}^\infty \frac{(-1)^n}{n!}q(t)^n\right) 
\end{multline}

\noindent The first order gives,

\begin{equation}
x(t)^{-1} = 1 + \frac{1-x_0}{x_0}e^{-\epsilon_\ell}\left[1 - \sum_{i=1}^N \frac{\alpha_i}{\omega_i}\sin(\varphi_i)\sin(\omega_i t - \varphi_i)\right].\label{eq:truncated_linear}
\end{equation}

\noindent The case with a single frequency can be compared with Eq.~\eqref{eq:freq_response0}.
The equivalence between the first order models requires that (see Supplementary data for comparison of responses),

\begin{gather}
\begin{aligned}
\lambda_\ell &= \epsilon\qquad \text{(delay)}\\
z_s &= -\ln\left(\frac{x_0}{1-x_0}\frac{\lambda}{\epsilon}\right) - \frac{m}{\lambda_\ell} \qquad \text{(mean value)}.
\end{aligned}\label{eq:params}
\end{gather}

\noindent This equalities introduce properties of the input into the system parameters.
Therefore, this Wiener model is input signal dependent.

As mentioned before, the fact that we can establish an equivalence (approximated) between the nonlinear model and the Wiener model indicates that the memory of the nonvolatile memristors will resemble that of a linear system.
To observe memories that differ from exponential decays, high amplitude oscillations will be required.
Note however, that the decay of conductance observed in experiments is not linear (see Fig.~\ref{fig:state_decay}), possibly allowing for more complicated memory functions in a wider range of amplitudes.

\section{Reservoir computing with memristors in series}
\label{sec:series}

In the RC context, the response of a set of memristors is used to assemble a design matrix which is used to linearly regress the desired output.
A simple but restrictive approach is to ask that the responses are linearly combined instantaneously,

\begin{equation}
\hat{y}_k = \vBQ (t_k)^{T} \bm{w}.
\end{equation} 

\noindent Where $t_k$ are the sampling timestamps, $\hat{y}_k$ is the approximation to the desired output $y_k$ and $\vBQ (t_k)$ is a vector with the responses of the set.
The mixing vector $\bm{w}$ is fixed for all $t_k$ and has as many components as there are memristors in the system.
For all timestamps we have

\begin{equation}
\bm{\hat{y}} = \vBQ \bm{w},
\end{equation} 

\noindent $\vBQ \in \mathbb{R}^{T \times n}$ is the design matrix having the $n$ responses of the set as columns. This approach will be used in section~\ref{sec:delay} to generate delayed versions of the input.

According to Eqs.~\eqref{eq:sine_w}-~\eqref{eq:cosine_w} the voltage drop across $n$ different memristors (different values of $\lambda$ and $\mu$) to a single frequency current of small amplitude (first order in $I_\omega$) can provide at most a design matrix of rank $2$ given by the sine and cosine components of the responses.
For an input with $N$ frequencies it is theoretically possible to obtain a maximum rank of $2N$, but probably this is unrealistic.
The use of larger amplitudes can increase the rank due to the generation of higher harmonics.
In this case, incommensurate frequencies in the input are the best option to maximize the rank, since they reduce the overlap of harmonics.
Interestingly, the rank benefits from memristor variability, making it a desired feature of the building process instead of a nuisance~\citep[][ and reference 98 therein]{Indiveri2013}.

\begin{figure}[htpb]
\centering
\includegraphics[scale=1]{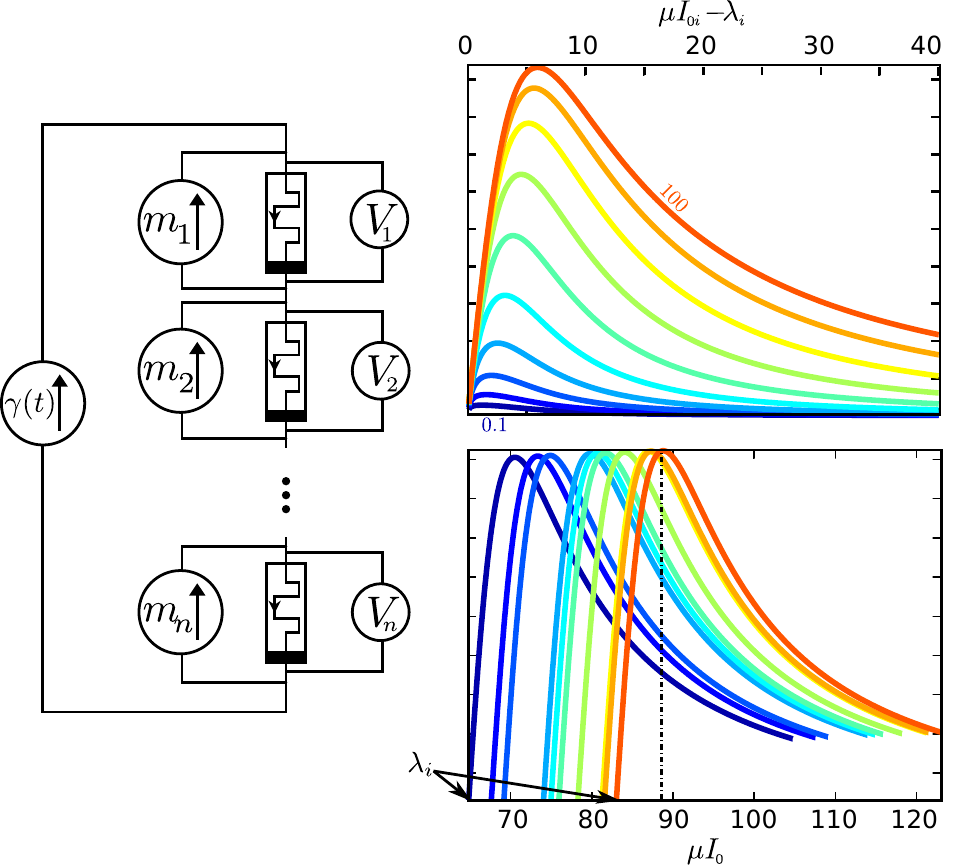}
\caption{Memristors in series.
(left) Set of memristors for simple signal processing.
Each memristors is fed a constant current $m_i = \mu_i I_{0i}$.
The right panel shows the cosine component (Eq.~\eqref{eq:cosine_w}, arbitrary scale) for independent $m_i$ (top) and for equal $m_i$ (bottom).
Labels on the curves indicate the value of $\lambda_i$.
Independent $m_i$ allow to tune $m_i - \lambda_i$ and the response of each memristor.
When a single mean value is present (dotted line in bottom panel) the response depends on the natural distribution of the parameters.}
\label{fig:series}
\end{figure}

Figure~\ref{fig:series} (left panel) shows a mockup setup of memristors connected in series.
The input current is a zero mean signal $\gamma(t)$ and each memristor is fed a small local current $I_{0i}$.
The case $I_{0i}=I_0$ for all $i$ corresponds to a signal with mean value and no local current.
The right panel of the figure shows the distribution of cosine components for independent mean values (top) and for a unique mean value (bottom).
For simplicity of the presentation, the bottom panel assumes that all memristors have the same $\mu$ parameter. 
In general the cosine contribution will be controlled by $\mu_i I_0 - \lambda_i$, wider variety will, in general, improve the rank of the design matrix.
In terms of control of the design matrix properties, the most versatile situation is the one with independent local inputs $m_i$.
When this is not possible, variability of the memristor parameters is the key for successful RC.
A wider variety responses to the same signal, i.e. parameter variability, would increase the rank of the design matrix.

Herein we assume that the diffusion parameter $\lambda$ and the effective ionic mobility $\mu$ are independent parameters.
The veracity of this assumption depends on the physical diffusive mechanism.
If diffusion is coupled with carrier mobility, then $\lambda$ will be proportional to $\mu$.
However, as in the case of the leaking hydraulic memristor this is not necessarily true in general.

\subsection{Delayer}
\label{sec:delay}
The delay task requires that the linear mixture of output voltages is able to recover a delayed version of the input signal. For the example presented here we have sampled $500$ input signals defined by,

\begin{equation}
\gamma(t) =  \alpha \sum_{n=1}^{12} \frac{\xi_n}{\pi n}\sin(\pi n t) - C.
\end{equation} 

\noindent With $\xi_n$ independent Gaussian variables with zero mean and unit variance.
The constant $C$ removes the mean value of the signal for $t \in [0,2]$.
The memristor bank consisted of $50$ units with $m_i - \lambda_i \in [0.1, 100]$ logarithmically spaced.
The response of the system was calculated for $36$ periods of the input.
The last $12$ periods were used for the training of the readout weights and a column filled with ones was added to the design matrix.
These weights were obtained using ridge regression with $10$-fold cross-validation.

\begin{figure}[htpb]
\centering
\includegraphics[width=0.9\textwidth]{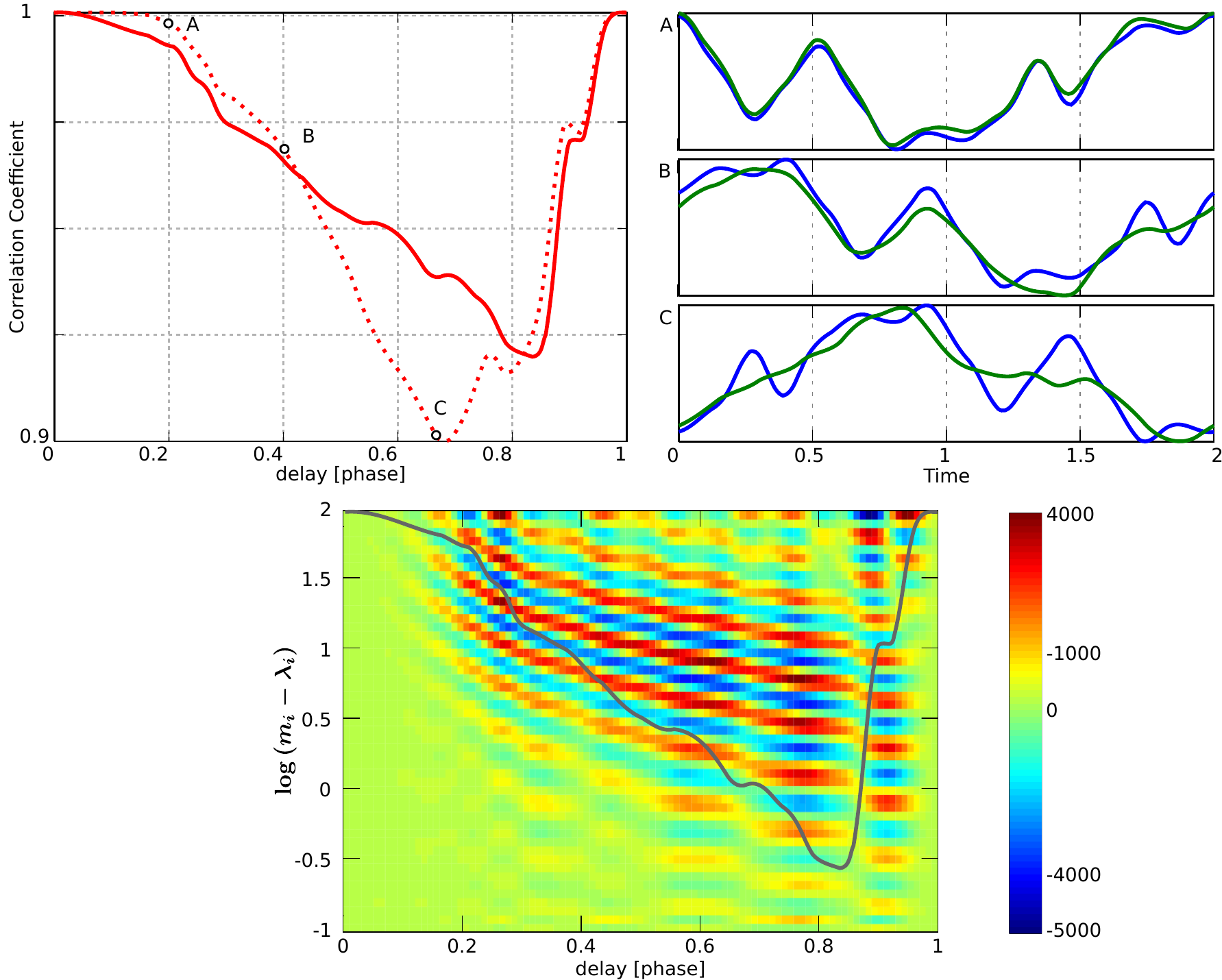}
\caption{Signal delayer.
Three panels showing the performance of the memristor bank for the delay task.
(top-left) Correlation coefficient between the estimated and desired output, for different delays.
The solid line corresponds to training values, the dotted line to a single test value.
(top-right) Test example for three delays (A,B,C marked in the panel) high-frequency features of the signal are lost with higher delays.
(bottom) Readout weights for each delay and $\epsilon_i$ parameter.
The correlation coefficient is overlaid to guide the eye.}
\label{fig:delay}
\end{figure}

The training performance of the system is shown in Fig.~\ref{fig:delay}, top-left panel.
The correlation coefficient interval is $[0.9, 1]$ indicating that the lower frequencies are more easily delayed (as indicated by Eq.~\eqref{eq:delay}).

The trained weights were tested using a previously unseen input signal shown in the top-right panel.
The deterioration of higher frequencies with increasing delay is also visible in these plots.

The panel at the bottom shows the distribution of the weights for each memristor (labeled with its $\epsilon_i$ parameter) and their variation with increasing delays.
Higher delays recruit memristors with smaller $\epsilon_i$.
The integrals in Eq.~\eqref{eq:taylor_resp} allows to interpret this in terms of memory, with lower $\epsilon_i$ corresponding to filters with longer time constants, i.e. longer memory.
Note that volatility cannot be cancel out with a fixed mean value, since even if $\epsilon_i = 0$, diffusion is still driven by the mean value of the input with a quadratic dependency on the state (Eq.~\eqref{eq:diff_nldrift}).
The smooth variation of the readout weights with the delay can be interpolated reducing the number of delays that need to be trained explicitly.

The numerical rank of the obtained design matrix in these experiments was approximately $5$, with a clear gap between the singular values beyond that rank (discrete rank-deficient).
Therefore, the size of memristor bank could be reduced.
The size of the bank in this example was chosen to show the smooth variation of the readout weights as a function of $\epsilon_i$.
Similar performances as the one shown here were obtained with banks consisting of $10$ memristors.

\subsection{Binary operators}
In this example we use data from the set $\mathbb{Z}^2/3 = \lbrace0,1,2\rbrace^2$ and we try to learn all mappings $\mathbb{Z}^2/3 \rightarrow \mathbb{Z}/3$, i.e. binary operators in $\mathbb{Z}/3$.
Note that in the absence of noise, a design matrix with rank $9$ would solve any of these tasks exactly.
To use the memristor bank with discrete datasets, we need an encoding from the domain of the dataset to the continuous input signal space. 
Each input pair $(s_1,s_2) \in \mathbb{Z}^2/3$ is mapped to an input signal via the formula,
\begin{equation}
  u(t,s_1,s_2) = \frac{s_1+1}{3} \sin\left(\omega_1 t\right) + \frac{s_2+1}{3} \cos\left(\omega_2 t\right),\label{eq:freqenc}
\end{equation}

\noindent where $\omega_1 = \nicefrac{\pi^2}{\sqrt{2}}$ and $\omega_2 = 3\pi\sqrt{3}$. 
These signals combine two (or more if desired) frequencies. The values $s_i$ equalize the driving signal by defining frequency components within $u(t,s_1,s_2)$.
Since we will be working with the long term response of the memristor bank, the ordering of the input pairs is irrelevant, i.e. the response of the system for symbols $i$ is independent of the response for symbols $i-1$.
This also means that we are not exploiting the memory of the dynamical system.
To do this, we would use encodings that present each element of the input pair sequentially.
The encoding~\eqref{eq:freqenc} allows us to use the bank as a static map between inputs and outputs similar to ELMs~\citep{Huang2004ELM}. The same operators can be implemented in nonvolatile memristors, granted that all encoded inputs have zero mean.

For this application, the readout map is a linear combination of buffered output signals:
for each input pair, we integrate the system for a period of time $T = \nicefrac{20\pi}{\omega_1}$ (sampling frequency $f = \unit{150}\hertz$) and sample the output voltages at timestamps $T - [0.95, 0.7, 0.5]\times \nicefrac{2\pi}{\omega_1}$.
With this data we construct a single row of the design matrix. 
Therefore, the total number of readout weights is $3N$, with $N=10$ the number of memristors in the bank.

\begin{figure}[tpb]
\centering
\includegraphics[scale=1]{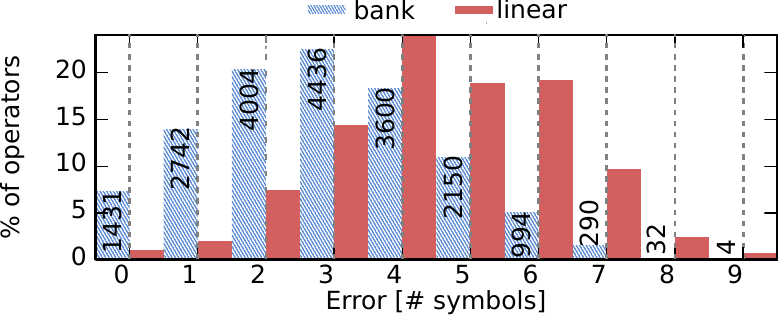}
\caption{Solving all tasks in $\mathbb{Z}/3$. 10-fold cross validation error for all possible tasks in $\mathbb{Z}/3$ for a single $10$ memristors bank. Each bar shows the tasks count and the y-axis shows the corresponding percentage. Pattern filled bars corresponds to the memristor bank, solid (red) bars are the results obtained using the inputs directly.}
\label{fig:z3all}
\end{figure}

Figure~\ref{fig:z3all} shows the performance of the memristor bank for all possible binary operators in $\mathbb{Z}/3$.
About $64 \%$ of all $3^{3^2}$ possible tasks are solved with $3$ or less errors.
For comparison, the plot also shows the performance of a design matrix constructed in the same way but using directly the input signals instead of the system's responses.

If we use random matrices as proxies for systems generating responses with very narrow autocorrelation functions, the performance of the memristor bank ($64 \%$ with $3$ or less errors) is obtained with random matrices of rank $7$ and above ($9\times30$ matrices with Gaussian entries and then SVD truncated to the desired rank).
However the design matrix generated by the memristor rank has a numerical rank of $3$ ($1\!\!\times\!\!10^{-2}$ tolerance).
This indicates that the singular vectors are better suited for the task than the ones from random matrices.

\section{Discussion}
We have studied two models of volatile memristors that are reasonable modifications of popular models used nowadays for the study of nonvolatile devices.
Model~\eqref{eq:nldrift} has been used for the simulation of networks~\citep{Stieg2014}, while model~\eqref{eq:linmodel} is an original contribution of the authors.
Interestingly the nonlinear model of memristors can be thought of as a static nonlinearity applied on the trajectories of a linear dynamics model, i.e. a Wiener model. 
Though the equivalence between the models could be used for nonlinear system identification, its existence might hinder the usability of these systems in reservoir computing, since memory functions are restricted to decaying exponentials.
A uniformly decaying memory function is unsuited for many machine learning problems, e.g. finding matching parentheses in a written sentence, which requires some kind of retrievable memory.
However, the volatility observed in real devices~\citep{Ohno2011} is more complex than the linear volatility studied herein.
Nonlinear volatility might prevent an equivalent Wiener model.
Additionally, memory can become intricate when the system is used in close loop. 

The addition of linear volatility to resistance switching models based on charge transport, implies that there is a flow of carriers that goes out of the electrical circuit of the device. 
Here in this additional sink was not explicitly modeled.
Improved models could include leak currents generated by diffusive processes and internal batteries, i.e. using Eqs.~\eqref{eq:extmem} and~\eqref{eq:extdyn}. 
Nevertheless, before moving towards more detailed models, we believe there is a need for thorough evaluation of the current models against the behavior of real devices.

When memristors are used in the RC framework, the natural variability of their parameters comes as a benefit contrary to what it is usually considered in other computational paradigms.
The variability of the devices produces a gamma of responses (ideally, linearly independent responses) to the same input signal.
This set can then be linearly combined to generate new functions.
In this regard, RC is similar to Galerkin's method but the set of linearly independent functions used as generators are provided by the system itself rather than by design.
This particularity makes the approach interesting for adaptive systems and has been used to generate motion control inputs for nonlinear systems~\citep[][part II]{CarbajalPhD2012} and to draw connections to the synergy hypothesis in motor control~\citep{Alessandro2013}.

Whether variations of the input can be afforded by the RC solution is highly dependent on the kind of system
generating the responses and must be studied on a case by case basis.
In general, systems showing fading memory for a family of driving signals are expected to be able to cope with input perturbations localized in time.
This point highlights the role of the input encoding.
The models studied here have a very well defined harmonic generation~\citep{Georgiou2012, Oskoee2011} and encoding information in the spectrum (e.g. Eq.~\eqref{eq:freqenc}) could exploit this natural behavior.
However these memristor models are suited to produce only upshifts of frequency bands in the lower range of the spectrum.
Another important feature of the model studied here is the coupling between the mean value of the inputs and the induced delay (and harmonic generation near the saturation).
This indicates that low frequency variations are well suited for encodings.
Note that this encoding would be much restricted in nonvolatile devices, since inputs with mean value would saturate the devices almost all the time.

The memristor bank presented in section~\ref{sec:series} is the simplest topology that can be assembled with a set of memristors.
It is a building block for the understanding of larger topologies.
It also offers a reference to evaluate the contribution of more complicated topologies.
To study the complementary circuit, the parallel topology, we need to adapt the methods in~\citep{Biolek2012} to include the time variant component of the dynamics.
This will be done in subsequent works.

Regarding numerical simulations of large networks with the two models presented herein, there are several issues that are worth mentioning.
For the case of the nonlinear state dynamics (Eq.~\eqref{eq:nldrift}), $x=0$ is unstable and any numerical error giving a slightly negative value will create a divergence of the simulations.
To solve this, the value of $x$ has to be forced to comply with $x>0$ (slowing down the simulations) or one can build a nonlinear equation that is stable in both extrema of the interval $[0,1]$.
When $\lambda = 0$, once the internal state reaches $x=0$ or $x=1$ beyond machine precision, it cannot be driven out of those states since it becomes insensitive to the driving signal.
This is an unavoidable problem when input signals have nonzero mean value.
This is not the case for the model in Eq.~\eqref{eq:linmodel}.
$H(x)$ can be driven out from saturation (in either extrema) using the appropriate input signal, even when numerical rounding errors
are present.
Note that model~\eqref{eq:nldrift} provides a bounded internal state, while~\eqref{eq:linmodel} does not.
Since in the latter model saturation is present in the output function but not in the internal states, the integral of the input needed to "de-saturate" the internal state may grow with time and input signals with mean value could drive the internal states to overflow.

\section{Conclusions}
In this work we have studied a modified version of the nonlinear Strukov memristor model that includes a linear diffusion term.
The long term behavior of the model and its response to periodic inputs have been presented in detail.
The results evinced the role of the mean value of the inputs as a modulator of the memory of the memristors and the consequent generation of delays.

We have also presented a Wiener model that can approximate the behavior of the nonlinear model for inputs with small amplitude variations. 
This model can provide a starting point for nonlinear system identification of large networks.
However, the existence of a Wiener model approximating the memristor behavior limits the complexity of its memory functions.
Feedback loops would be required to overcome this drawback. 
 
We have presented results on two academic tasks: delaying an input signal and binary operators in $\mathbb{Z}/3$.
The delay task illustrated how RC recruits memristors with longer memories to generate higher delays.
The binary operator task, showed that although the design matrices are rank-deficient their performance is comparable to high rank random matrices.
The results also highlight that device variability is a desired feature for the implementation of RC in memristor networks.

Clearly, the different computational properties arising from the Strukov-based model and the Wiener model imply that a more in-depth study on the use of memristor networks for reservoir computing relies on the selection of a physically realistic model.
Therefore, experimental modeling research is required for the development of accurate volatile memristor models.

%%%%%%%%%%%%%%%%%%%%%%%%%%%%%%%%%%%%%%%%%%%%%%%%%%%%%%%%%%%%%%%%%%%%%%%%%%%%%%%%
\section*{Acknowledgements}
%%%%%%%%%%%%%%%%%%%%%%%%%%%%%%%%%%%%%%%%%%%%%%%%%%%%%%%%%%%%%%%%%%%%%%%%%%%%%%%%

The authors would like to thank Prof. Nir Y. Krakauer for the suggested literature on rank-deficient linear problems. We acknowledge the fruitful discussions with Dr. Pieter Buteneers about ELMs. We thank the developers of~\citetalias{Octave, Sage} and~\citetalias{Inkscape} for their excellent software tools.

\paragraph*{Funding.} The research leading to these results has received funding from the European Union Seventh Framework Programme (FP7/2007-2013) under grant agreement no. 604102 (Human Brain Project).

\paragraph{Author contributions}: \textbf{JPC} developed the software and mathematical formulation; carried out the experiments, data analysis. \textbf{JPC} \& \textbf{JD} wrote this manuscript. \textbf{BS} \& \textbf{MH} contributed to the numerical experiment design and copy-edition.

\bibliographystyle{vancouver}
\bibliography{Memristors}

\begin{thebibliography}{10}

\bibitem{Indiveri2013}
Indiveri G, Linares-Barranco B, Legenstein R, Deligeorgis G, Prodromakis T.
\newblock {Integration of nanoscale memristor synapses in neuromorphic
  computing architectures.}
\newblock Nanotechnology. 2013 Sep;24(38):384010.

\bibitem{Kuzum2013}
Kuzum D, Yu S, Wong HSP.
\newblock {Synaptic electronics: materials, devices and applications.}
\newblock Nanotechnology. 2013 Sep;24(38):382001.

\bibitem{Esmaeilzadeh2012}
Esmaeilzadeh H, Sampson A, Ceze L, Burger D.
\newblock {Neural Acceleration for General-Purpose Approximate Programs}.
\newblock In: 45th Annual IEEE/ACM International Symposium on Microarchitecture
  (MICRO). IEEE; 2012. p. 449 -- 460.

\bibitem{Samadi2013}
Samadi M, Lee J, Jamshidi DA, Hormati A, Mahlke S.
\newblock {SAGE: self-tuning approximation for graphics engines}.
\newblock In: 46th Annual IEEE/ACM International Symposium on Microarchitecture
  (MICRO). IEEE; 2013. p. 13--24.

\bibitem{Maass2002}
Maass W, Natschl\"{a}ger T, Markram H.
\newblock {Real-time computing without stable states: a new framework for
  neural computation based on perturbations.}
\newblock Neural computation. 2002 Nov;14(11):2531--60.

\bibitem{Jaeger2004}
Jaeger H, Haas H.
\newblock {Harnessing nonlinearity: predicting chaotic systems and saving
  energy in wireless communication.}
\newblock Science (New York, NY). 2004 Apr;304(5667):78--80.

\bibitem{Maass10}
Maass W.
\newblock {Liquid state machines: motivation, theory, and applications}.
\newblock In: Cooper SB, Sorbi A, editors. In Computability in Context:
  Computation and Logic in \ldots. Imperial College Press; 2010. p. 275--296.

\bibitem{Lukosevicius2012}
Luko\v{s}evi\v{c}ius M, Jaeger H, Schrauwen B.
\newblock {Reservoir Computing Trends}.
\newblock KI - K\"{u}nstliche Intelligenz. 2012 May;26(4):365--371.

\bibitem{Triefenbach10}
Triefenbach F, Jalalvand A, Schrauwen B, Martens JP.
\newblock {Phoneme recognition with large hierarchical reservoirs}.
\newblock In: Lafferty J, Williams CKI, Shawe-Taylor J, Zemel RS, Culotta A,
  editors. Advances in Neural Information Processing Systems. vol.~23. Neural
  Information Processing System Foundation; 2010. p.~9.

\bibitem{Buteneers2013}
Buteneers P, Verstraeten D, Van~Nieuwenhuyse B, Stroobandt D, Raedt R, Vonck K,
  et~al.
\newblock Real-time detection of epileptic seizures in animal models using
  reservoir computing.
\newblock EPILEPSY RESEARCH. 2013;103(2-3):124--134.

\bibitem{Ongenae2013}
Ongenae F, Van~Looy S, Verstraeten D, Verplancke T, Benoit D, De~Turck F,
  et~al.
\newblock Time series classification for the prediction of dialysis in
  critically ill patients using echo state networks.
\newblock ENGINEERING APPLICATIONS OF ARTIFICIAL INTELLIGENCE.
  2013;26(3):984--996.

\bibitem{Alessandro2013}
Alessandro C, Carbajal JP, D'Avella A.
\newblock {A computational analysis of motor synergies by dynamic response
  decomposition}.
\newblock Front Comput Neurosci. 2013;.

\bibitem{Sillin2013}
Sillin HO, Aguilera R, Shieh HH, Avizienis AV, Aono M, Stieg AZ, et~al.
\newblock {A theoretical and experimental study of neuromorphic atomic switch
  networks for reservoir computing.}
\newblock Nanotechnology. 2013 Sep;24(38):384004.

\bibitem{Reinhart2012}
Reinhart RF, {Jakob Steil} J.
\newblock {Regularization and stability in reservoir networks with output
  feedback}.
\newblock Neurocomputing. 2012 Aug;90:96--105.

\bibitem{Fiers2014}
Fiers M, Van~Vaerenbergh T, wyffels F, Verstraeten D, Schrauwen B, Dambre J,
  et~al.
\newblock Nanophotonic reservoir computing with photonic crystal cavities to
  generate periodic patterns.
\newblock IEEE TRANSACTIONS ON NEURAL NETWORKS AND LEARNING SYSTEMS.
  2014;25(2):344--355.

\bibitem{wyffels2014}
wyffels F, Li J, Waegeman T, Schrauwen B, Jaeger H.
\newblock Frequency modulation of large oscillatory neural networks.
\newblock BIOLOGICAL CYBERNETICS. 2014;.

\bibitem{Fernando2003}
Fernando C, Sojakka S.
\newblock {Pattern Recognition in a Bucket}.
\newblock In: Banzhaf W, Ziegler J, Christaller T, Dittrich P, Kim JT, editors.
  Advances in Artificial Life. Springer Berlin Heidelberg; 2003. p. 588--597.

\bibitem{Caluwaerts2013}
Caluwaerts K, D'Haene M, Verstraeten D, Schrauwen B.
\newblock Locomotion without a brain: physical reservoir computing in
  tensegrity structures.
\newblock ARTIFICIAL LIFE. 2013;19(1).

\bibitem{Vandoorne2014}
Vandoorne K, Mechet P, Van~Vaerenbergh T, Fiers M, Morthier G, Verstraeten D,
  et~al.
\newblock {Experimental demonstration of reservoir computing on a silicon
  photonics chip}.
\newblock Nature Communications. 2014;5.

\bibitem{Paquot2012}
Paquot Y, Duport F, Smerieri A, Dambre J, Schrauwen B, Haelterman M, et~al.
\newblock {Optoelectronic reservoir computing}.
\newblock Scientific Reports. 2012;2:1--6.

\bibitem{Larger2012}
Larger L, Soriano MC, Brunner D, Appeltant L, Gutierrez JM, Pesquera L, et~al.
\newblock {Photonic information processing beyond Turing: an optoelectronic
  implementation of reservoir computing}.
\newblock Optics Express. 2012;20(3):3241.

\bibitem{Hermans2010}
Hermans M, Schrauwen B.
\newblock {Memory in linear recurrent neural networks in continuous time.}
\newblock Neural networks : the official journal of the International Neural
  Network Society. 2010 Apr;23(3):341--55.

\bibitem{Manjunath13}
Manjunath G, Jaeger H.
\newblock {Echo state property linked to an input: exploring a fundamental
  characteristic of recurrent neural networks.}
\newblock Neural computation. 2013 Mar;25(3):671--96.

\bibitem{MacLennan2012}
MacLennan BJ.
\newblock {Analog Computation}.
\newblock In: Meyers RA, editor. Computational Complexity. New York: Springer;
  2012. p. 161--184.

\bibitem{Hansen1998}
Hansen PC.
\newblock {Rank-Deficient and Discrete Ill-Posed Problems}.
\newblock Society for Industrial and Applied Mathematics; 1998.

\bibitem{Dambre2012}
Dambre J, Verstraeten D, Schrauwen B, Massar S.
\newblock {Information processing capacity of dynamical systems.}
\newblock Scientific reports. 2012 Jan;2:514.

\bibitem{Vongehr2012}
Vongehr S.
\newblock {Missing the Memristor}.
\newblock Advanced Science Letters. 2012 Oct;17(1):285--290.

\bibitem{Avizienis2012}
Avizienis AV, Sillin HO, Martin-Olmos C, Shieh HH, Aono M, Stieg AZ, et~al.
\newblock {Neuromorphic atomic switch networks.}
\newblock PloS one. 2012 Jan;7(8):e42772.

\bibitem{Stieg2012}
Stieg AZ, Avizienis AV, Sillin HO, Martin-olmos C, Aono M, Gimzewski JK.
\newblock {Emergent Criticality in Complex Turing B-Type Atomic Switch
  Networks}.
\newblock Advanced Materials. 2012;24(2):286--293.

\bibitem{Ohno2011}
Ohno T, Hasegawa T, Nayak A, Tsuruoka T, Gimzewski JK, Aono M.
\newblock {Sensory and short-term memory formations observed in a Ag2S gap-type
  atomic switch}.
\newblock Applied Physics Letters. 2011;99(20):203108.

\bibitem{Strukov08}
Strukov DB, Snider GS, Stewart DR, Williams RS.
\newblock {The missing memristor found.}
\newblock Nature. 2008;453(7191):80--83.

\bibitem{Chua1971}
Chua LO.
\newblock {Memristor-the missing circuit element}.
\newblock IEEE Transactions on Circuit Theory. 1971;ct-18(5):507--519.

\bibitem{Valov2013}
Valov I, Linn E, Tappertzhofen S, Schmelzer S, van~den Hurk J, Lentz F, et~al.
\newblock {Nanobatteries in redox-based resistive switches require extension of
  memristor theory}.
\newblock Nature Communications. 2013 Apr;4:1771.

\bibitem{Biolek2012}
Biolek Z, Biolek D, Biolkova V.
\newblock {Analytical Solution of Circuits Employing Voltage- and
  Current-Excited Memristors}.
\newblock IEEE Transactions on Circuits and Systems I: Regular Papers. 2012
  Nov;59(11):2619--2628.

\bibitem{Oskoee2011}
{Nedaaee Oskoee} E, Sahimi M.
\newblock {Electric currents in networks of interconnected memristors}.
\newblock Physical Review E. 2011 Mar;83(3):031105.

\bibitem{Huang2004ELM}
Huang GB, Zhu QY, Siew CK.
\newblock {Extreme learning machine: a new learning scheme of feedforward
  neural networks}.
\newblock In: Neural Networks, 2004. Proceedings. 2004 IEEE International Joint
  Conference on. vol.~2. IEEE; 2004. p. 985--990.

\bibitem{Stieg2014}
Stieg AZ, Avizienis AV, Sillin HO, Aguilera R, Shieh Hh, Martin-olmos C, et~al.
\newblock {Memristor Networks}.
\newblock Adamatzky A, Chua L, editors. Cham: Springer International
  Publishing; 2014.

\bibitem{CarbajalPhD2012}
Carbajal JP.
\newblock {Harnessing Nonlinearities: Generating Behavior from Natural
  Dynamics} [PhD].
\newblock University of Z\"{u}rich; 2012.

\bibitem{Georgiou2012}
Georgiou PS, Barahona M, Yaliraki SN, Drakakis EM.
\newblock {Device Properties of Bernoulli Memristors}.
\newblock Proceedings of the IEEE. 2012 Jun;100(6):1938--1950.

\bibitem{Octave}
{Octave community}. {GNU Octave} 3.8.1; 2014.
\newblock Available from: \url{www.gnu.org/software/octave/}.

\bibitem{Sage}
{The Sage Development Team}. {S}age {M}athematics {S}oftware ({V}ersion 6.1.1);
  2014.
\newblock Available from: \url{http://www.sagemath.org}.

\bibitem{Inkscape}
{Inkscape community}. {I}nkscape 0.48; 2014.
\newblock Available from: \url{http://www.inkscape.org/}.

\end{thebibliography}

\end{document}